\documentclass[12pt]{article}

 \usepackage{authblk}

 \usepackage{amsfonts}
 \usepackage{amsmath}
 \usepackage{amssymb}
 \usepackage{amsthm}

 \usepackage{graphicx}
 \usepackage[update,prepend]{epstopdf}

 \usepackage{booktabs} % For formal tables

 \usepackage{caption}

 \usepackage{multirow}

 \usepackage{float}

   \newtheorem*{remark}{Remark}
   \DeclareMathOperator*{\argmin}{arg\,min}
   \DeclareMathOperator*{\argmax}{arg\,max}
   \DeclareMathOperator*{\trace}{trace}
   \DeclareMathOperator*{\diag}{diag}

\title{Maximum Margin Principal Components}
\author[1]{Xianghui Luo\thanks{xluo@xaut.edu.cn}}
\author[2]{Robert J. Durrant\thanks{bobd@waikato.ac.nz}}
\affil[1]{Department of Computer Science}
\affil[2]{Department of Mathematics and Statistics}
\affil[ ]{University of Waikato}
\affil[ ]{Hamilton, New Zealand}

\date{} % delete this line to display the current date

\begin{document}

\maketitle

\begin{abstract}
\textit{Principal Component Analysis (PCA)} is a very successful dimensionality reduction technique, widely used in predictive modeling. A key factor in its widespread use in this domain is the fact that the projection of a dataset onto its first $K$ principal components minimizes the sum of squared errors between the original data and the projected data over all possible rank $K$ projections. Thus, PCA provides optimal low-rank representations of data for least-squares linear regression under standard modeling assumptions. On the other hand, when the loss function for a prediction problem is not the least-squares error, PCA is typically a heuristic choice of dimensionality reduction -- in particular for classification problems under the zero-one loss. In this paper we target classification problems by proposing a straightforward alternative to PCA that aims to minimize the difference in \textit{margin distribution} between the original and the projected data. Extensive experiments show that our simple approach typically outperforms PCA on any particular dataset, in terms of classification error, though this difference is not always statistically significant, and despite being a filter method is frequently competitive with Partial Least Squares (PLS) and Lasso on a wide range of datasets.
\end{abstract}

\section{Background and Introduction}

Dimensionality reduction techniques are a core part of the Statistics and Machine Learning toolbox, widely used in predictive modeling to improve a range of measures including generalization performance, interpretability or identifiability of models, and the time and space complexity of learning or prediction. There are many such techniques, see e.g. \cite{fodor2002survey} for a survey, but chief amongst them are simple linear techniques and the most widely-used of these in practice is probably Principal Components Analysis (PCA) \cite{Jolliffe2002a} and its variants. Applications of PCA include \cite{turk1991eigenfaces,turk1991face,Zhao2003}.

PCA works as follows: Suppose we have a data matrix $X \in \mathbb{R}^{D \times N}$ of $N$, $D$-dimensional observations. PCA works by linearly projecting our original $D$-dimensional data  onto $K$ uncorrelated (orthogonal) directions -- the first $K$ `Principal Components' -- where typically $K \ll D$. Denote by $P \in\mathbb{R}^{D \times K}$ the matrix with the Principal Components as columns, then the Principal Components are chosen to maximize the orthogonal projection of the dataset $X$ onto the column space of $P$, that is P is chosen to satisfy:
\begin{equation}
\min_{P} \|X - PP^{T}X\|^{2}_{2}.
\end{equation}
As a result, PCA gives the best $K$-dimensional representation of the original $D$-dimensional data in the least-squares sense or, equivalently, if the data are centered, using PCA means that (for a fixed dataset) we discard the smallest amount of the total sample variation of any linear dimensionality reduction scheme. For data analysis tasks other than Ordinary Least Squares Regression (OLS) using PCA is a heuristic which works well frequently but, as noted in \cite{haufe2014interpretation, Pechenizkiy2004, Jolliffe1982, Hill1977}, it can work very poorly (even for a linear regression task) in some natural scenarios. For the task of classification, PCA often seems to work very well experimentally \cite{howley2006effect,janecek2008relationship}, but it is trivial to construct examples for which PCA will work badly in a classification task\footnote{For example, when the most discriminative features have small between-class variance relative to the variance of the whole dataset, or relative to the within-class variance.}. In other words since the PCA objective is disconnected from the classification task at hand and, in particular it does not take account of the class structure inherent in the problem, vanilla PCA is prone to underfitting. In supervised settings where class labels are present, it would be a waste not to use the label information in the selection of useful features for classification, thus we propose a (weakly) supervised variant of PCA. There are a few supervised dimensionality reduction approaches, of which the most related to PCA is Fisher Linear Discriminant (FLD). FLD picks out the most useful feature for classification by maximizing the ratio of between-class variance to with-in class variance. It is supposed to outperform PCA for classification tasks in theory. However, as pointed out in \cite{Martinez2001}, PCA often outperforms FLD in practice, especially for small datasets. This is due the poor estimations of between-class variance and with-in class variance. Another important and widely used supervised approach is the wrapper methods based on partial least squares (PLS), e.g. \cite{brereton2014partial} and references therein. These take advantage of known structure in the classification problem by rotating the uncorrelated components to minimize the least square error on the predicted labels. While such wrapper approaches are frequently successful in practice, it is well known that they need careful tuning to avoid overfitting especially when sample size is small \cite{brereton2014partial}. More recently, a discriminative feature extraction method inspired by PCA was proposed in \cite{Karampatziakis2014}. This method picks out the directions along which two classes differ most in their class-conditional second moments. However, it also suffers from the inaccurate second moment estimation problem that plagues FLD in the case of small size high dimensional datasets. Finally, the $\ell_1$ regularization and sparse model selection method \textit{Lasso} \cite{Tibshirani1996} widely used in regression can also be used for dimensionality reduction for classification tasks, like $\ell_1$-regularized SVM. LARS \cite{Efron2004} is another similar approach.

In this paper we describe and empirically validate a PCA variant that attempts to address both issues: In particular, we choose a projection that maximally preserves (a proxy for) the sample margin distribution between two classes. Our approach is very simple to implement and only involves changing several lines of code to vanilla PCA but, as far as we are aware, it is neither previously published nor folklore. More importantly it has the same time and space complexity as PCA, targets a sensible objective (for classification) yet it is not a wrapper method, has a clear interpretation in terms of the problem at hand, and in our experience it works unreasonably well -- our experimental outcomes are typically better than vanilla PCA, frequently significantly better, and never significantly worse, and also competitive with PLS and Lasso.

The remainder of this paper is structured as follows: In the next section we describe PCA more precisely and recall that the principal components are the solution to a straightforward eigenvector problem. Next we discuss the importance of the \textit{margin distribution} as a measure of the difficulty of a classification problem, which motivates our altering the PCA objective in order to maximally preserve the sample margin distribution. Then we introduce several schemes based on our idea and provide some theoretical intuitions for how they work. We present extensive experiments on several datasets which demonstrate the utility of our approach and compare its performance to vanilla PCA, PLS, and Lasso, we use a simple non-parametric sign test to demonstrate the statistically significant superiority of our approach over vanilla PCA. Finally we summarize our findings and discuss some possible future directions for this research.

\section{Preliminaries}

Our main contribution is a novel PCA variant that aims to preserve the margin distribution, and in this section we therefore briefly review vanilla PCA, minimum margin and the margin distribution.

\subsection{Principal Component Analysis}

Given $N$ points $\{x_1,x_2,\cdots,x_N\}$ in $\mathbb{R}^D$ and a target dimension $K \leq D$, PCA finds a linear projection $P: \mathbb{R}^D \to \mathbb{R}^K$ and embedding $R: \mathbb{R}^K \to \mathbb{R}^D$ such that
\begin{equation}
 \sum_{i=1}^N \| x_i - R(P(x_i)) \|_2^2
\end{equation}
is minimized. In words, PCA finds the linear projection $P$ that best preserves the data in the least squares sense when the reconstruction $R$ is also a linear map. Representing $P$ and $R$ by matrices, it is easy to show that the optimal solution satisfies:
\begin{equation}
R = P^T.
\end{equation}
Therefore, PCA amounts to solving the convex optimization problem:
\begin{equation}
\argmin_{P \in \mathbb{R}^{K \times D}} \sum_{i=1}^N \| x_i - P^{T}Px_i \|_2^2,
\label{eq:pca_opt}
\end{equation}
where $\mathbb{R}^{K \times D}$ is the space of $K\times D$ matrices. Solving (\ref{eq:pca_opt}) is equivalent to solving:
\begin{equation}
\argmax_{P \in \mathbb{R}^{K \times D}:PP^T=I_K} \trace(P A P^T),
\end{equation}
where $A = \sum_{i=1}^N x_i x_i^T$. Since $A$ is positive semi-definite the principal components can be found simply by diagonalizing $A$ as $A = U^T \Lambda U$ where $U$ is orthogonal, $U^T U = UU^T = I_D$, and $\Lambda$ is a diagonal matrix of the non-negative eigenvalues of $A$ in descending order of magnitude. The first $K$ principle components comprise the eigenvectors of $A$ corresponding to the $K$ largest eigenvalues: That is, the first $K$ rows of $U$.

The computational cost of PCA is dominated by constructing the matrix $A$ and diagonalizing it. The former step takes $O(ND^2)$ calculations and the latter costs $O(D^3)$, though if $N < D$ the time complexity can be cut to $O(N^3)+O(DN^2)$. Thus the overall time complexity for PCA is $\min \{O(ND^2)+O(D^3), O(N^3)+O(DN^2)\}$ depending on whether the sample size is greater than the dimensionality or not.
\begin{remark}
Above we introduced PCA as a linear (lossy) compression approach. The well-known PCA as a dimensionality reduction approach is actually to apply the common practice to run it on the `mean-centered' data. Denote by $\hat{\mu} := \frac{1}{N}\sum_{i=1}^N x_i$ and take $A := \sum_{i=1}^N (x_i - \hat{\mu})(x_i - \hat{\mu})^T$, that is $A$ is proportional to the maximum likelihood consistent estimator of the covariance matrix. In this case the projection onto the column space of P maximizes the retained total sample variance. For some further perspectives on PCA, we refer the interested reader to \cite{Bie2005}.
\end{remark}

\subsection{Margin and Margin Distribution}

Let $\mathcal{H}$ be a hypothesis class of linear classifiers (separating hyperplane normals). The \textit{minimum margin} or just \textit{margin} between two separable classes is defined to be the supremum of the smallest Euclidean distances between the members of the classes and the separating hyperplane of classifiers $h \in \mathcal{H}$. That is, if $\mathcal{D}_-,\mathcal{D}_+$ represent the distribution of the two separable classes with $\mathcal{D} = \mathcal{D}_- \cup \mathcal{D}_+$, then we define the margin $\gamma$ to be:
\begin{equation}
 \gamma := \sup_{h \in \mathcal{H}}\inf_{x \in \mathcal{D}} y(x) h^T(x-z), 
\end{equation}
where $z$ is any point on the separating hyperplane, $y(x) \in \{-1,+1\}$ is the label of $x$, $\sup$ and $\inf$ are the supremum and infimum operators. For non-separable classes the margin is negative\footnote{Our definition of margin here is different from conventional definitions.}, so one can either enforce a non-zero `soft' margin by allowing a fraction of observations to be misclassified at training, or given a fixed classifying hyperplane consider instead the distribution of the signed distances between the members of classes and the fixed classifying hyperplane by defining the margin at the point $x \in \mathcal{D}$ with respect to classifier $h$ by:
\begin{equation}
\gamma(x,h) := y(x) h^T(x-z),
\end{equation}
where again $z$ is any point on the separating hyperplane. We call the distribution of $\gamma(x,h) := y(x) h^T(x-z),\, \forall x \in \mathcal{D}$ the \textit{margin distribution} with respect to $h$. The empirical margin and empirical margin distribution of a sample are defined similarly -- for a fixed classifier $\hat{h} \in \mathcal{H}$ and training set $\mathcal{D}_N$ of size $N$, we define the empirical margin:
\begin{equation}
\hat{\gamma} := \min_{x \in \mathcal{D}_N} y(x) \hat{h}^T(x-z),
\end{equation}
where $z$ represents a point on the separating hyperplane of $\hat{h}$, and the empirical margin at a point $x_i \in \mathcal{D}_N$ with respect to $\hat{h}$ as:
\begin{equation}\label{eq-margin}
\hat{\gamma}(x_i, \hat{h}) := y(x_i) \hat{h}^T(x_i-z)
\end{equation}
The empirical margin distribution with respect to $\hat{h}$ is defined as the distribution of $\hat{\gamma}(x_i, \hat{h}),\, \forall x_i \in \mathcal{D}_N$.

The importance of the margin between classes for classification has been studied extensively, where it can be viewed as a dimension-free measure of the difficulty of a classification task. For separable classes, \cite{Vapnik2000} bounds the generalization performance of the support vector machine (SVM) in terms of the empirical margin of the training dataset, and indeed both hard- and soft-margin SVM learn a discriminative hyperplane which maximizes the single-point minimum margin \cite{Zhang2016}. It has been pointed out that the information of the margin distribution is largely lost in the minimum margin which depends so critically on a small number of the training data points \cite{Shawe-Taylor1999}, but there is no such problem for the margin distribution. Thus tighter bounds for the generalization error of classifiers utilizing the margin distribution as the measure of difficulty of a classification problem have been obtained in, for example: \cite{Freund1999,Shawe-Taylor1999a,Shawe-Taylor1999}. Alternative classification algorithms based on the idea of margin distribution optimization have also been proposed \cite{Garg2003,Pelckmans2008,Aiolli2008,Zhang2014,Zhang2016} and found empirically to outperform SVM for many real datasets. Therefore, the margin distribution provides discriminative information that is crucial for classification. However, the margin and margin distribution can only be evaluated once the classifier has been learned. The dimensionality reduction schemes we introduce in this paper are based on the idea of preserving the margin distribution, but crucially without having access to the classifier.

\section{Algorithms}

\subsection{Motivation}

From the definitions above it is clear that optimizing the margin distribution requires a classifying hyperplane from which to measure it, and applying PCA to preserve exactly the sample margin distribution would therefore require a wrapper approach which could be prone to overfitting in small sample conditions. We therefore propose to run PCA on a proxy for the optimal margin distribution to obtain uncorrelated features that approximately preserve the margin distribution, and hence the important discriminative information for classification, without the same risk of overfitting.

Our heuristic argument runs as follows: PCA is the best linear (lossy) compression method if the reconstruction process is also linear. Therefore, the strength of PCA is in preserving the data on which it is applied to. For the purpose of dimensionality reduction for linear classification tasks, we are not interested in preserving the data points, but we are interested in selecting the features best for discriminating the data points. If we know in advance what information is useful for classification, we can extract some structures containing that information from the dataset and run PCA on those structures to obtain the features that are best for preserving them. If we then reduce the dimension of the original data by projecting to these features, we expect that the discriminative information is preserved in a good way by the projection. Based on this intuition, we devised four PCA variants for the task of linear classification, which we will evaluate in Section \ref{sec-theory}, \ref{sec-exp}. For now, we simply present the four heuristic alternatives in this section.

To illustrate the basic ideas, we focus on the two-class classification problem since generalization to  multi-class cases is straightforward. Let $\{(x_i,y_i)\}_{i=1}^N$ be a set of labeled training data points, where for convenience we assume $x_i$ is a point in $\mathbb{R}^D$, and $y_i \in \{-1,+1\}$ is the class label, $\forall i \in \{1,\cdots,N\}$. The crucial question is how to approximate the margin distribution of the dataset without access to a classifying hyperplane. Now the margin distribution contains the information about the differences between the data points of the two classes that is meaningful for the classification problem. Therefore, our first PCA variant represents the margin distribution as the differences between the data points of the two classes. Let $C_-$ and $C_+$ be the sets of indices of the data points that belong to class -1 and class +1 respectively, i.e. $C_- = \{i: y_i = -1\}, C_+ = \{j: y_j = +1\}$. 

\subsection{Algorithm: M-PCA0}

Define the following structures
\begin{equation}
z_{ij} := x_i - x_j,\quad \forall i \in C_-,\, j \in C_+.
\end{equation}
Then the structures $\{z_{ij}\}_{i\in C_-,j\in C_+}$ should reasonably represent the margin distribution well. We can then run ``uncentered'' PCA on these structures to obtain the most significant principal components. These principal components are the features that best represent this proxy for margin distribution and therefore contain the most discriminative information. We call this scheme \textit{M-PCA0}. The algorithm of M-PCA0 is shown in Figure~\ref{figure-mpca0}.

\begin{figure}[h]
\begin{center}
\begin{tabular}{|ll|}
\hline
  \multicolumn{2}{|c|}{\textbf{M-PCA0}}\\
  \textbf{input}:& $X=(x_1,\cdots,x_N)^T$ \\
   & {$Y=(y_1,\cdots,y_N)^T$}\\
   & {target dimension $K$}\\
  \multicolumn{2}{|l|}{let $C_-=\{i:y_i=-1\},\, C_+=\{i:y_i=+1\}$}\\
  \multicolumn{2}{|l|}{let $z_{ij} = x_i - x_j,\, \forall i \in C_-, j \in C_+$}\\
  \multicolumn{2}{|l|}{let $A=\sum_{i \in C_-, j \in C_+} z_{ij} z_{ij}^T$}\\
  \multicolumn{2}{|l|}{let $v_1,\cdots,v_K$ be the eigenvectors of $A$}\\
  \multicolumn{2}{|l|}{$\quad$ with the largest eigenvalues}\\
  \multicolumn{2}{|l|}{let $P = [v_1,\cdots,v_K]$}\\
  \textbf{output}:& $\tilde{X} = XP$, $\tilde{Y} = Y$\\
\hline
\end{tabular}
\caption{The M-PCA0 Algorithm}\label{figure-mpca0}
\end{center}
\end{figure}

It is not hard to see that the size of $\{z_{ij}\}_{i\in C_-,j\in C_+}$ is $N_-N_+$, where $N_- = | C_- |$, $N_+ = | C_+ |$, which is very large for large sample data. As a result, the M-PCA0 algorithm is more computationally expensive than usual PCA, it will be $\min \{O(N_-N_+D^2)+O(D^3), O(N_-^3N_+^3)+O(DN_-^2N_+^2)\}$.

\subsection{Algorithm: M-PCA1a and M-PCA1b}

With a little more consideration, it is not hard to come up with a natural alternative to $\{z_{ij}\}_{i\in C_-,j\in C_+}$ that resolves this time complexity issue. The Maximum Likelihood (ML) consistent estimates of the class means are
\begin{equation}
\hat{\mu}_- = \frac{1}{N_-} \sum_{i\in C_-} x_i,\quad \hat{\mu}_+ = \frac{1}{N_+} \sum_{j \in C_+} x_j.
\end{equation}
Define the variables
\begin{equation}\label{z_i}
z_k := \left\{ 
              \begin{aligned}
                x_k - \hat{\mu}_+\quad & \text{if $k\in C_-$}\\
                \hat{\mu}_- - x_k\quad & \text{if $k\in C_+$}
              \end{aligned}
 \right. ,\, \forall k \in \{1,\cdots,N\}
\end{equation}
In words, $z_k$ is defined as the difference between $x_k$ and the mean of the other class. Intuitively, $\{z_k\}_{k=1}^N$ captures the same information as $z_{ij}$ and it can be viewed as a conditioning of the previous problem. The sample size is now $N$ which is typically much smaller than the size of $\{z_{ij}\}_{i\in C_-,j\in C_+}$. We call the resulting algorithm  \textit{M-PCA1a}.

In the definition (\ref{z_i}), the sample means of the two classes are used. Especially in the situation of small sample data or very imbalanced classes, the quality of the estimates of the class conditional means is a concern. In this case, replacing the sample means with the sample medoids may be a more robust option\footnote{The class-conditional medoid is the vector consisting of the class-conditional sample median of each individual feature.}. We call the corresponding algorithm \textit{M-PCA1b}. Both versions of the algorithm are shown in  Figure~\ref{figure-mpca1}.  Time complexity for this approach is the same as the standard PCA.
\begin{figure}[!h]
\begin{center}
\begin{tabular}{|ll|}
\hline
  \multicolumn{2}{|c|}{\textbf{M-PCA1a(M-PCA1b)}}\\
  \textbf{input}:& $X=(x_1,\cdots,x_N)^T$ \\
   & {$Y=(y_1,\cdots,y_N)^T$}\\
   & {target dimension $K$}\\
  \multicolumn{2}{|l|}{let $C_-=\{i:y_i=-1\},\, C_+=\{i:y_i=+1\}$}\\
  \multicolumn{2}{|l|}{let $\hat{\mu}_-$ be the mean(medoid) of $\{x_i:i \in C_-\}$}\\
  \multicolumn{2}{|l|}{let $\hat{\mu}_+$ be the mean(medoid) of $\{x_j:j \in C_+\}$}\\
  \multicolumn{2}{|l|}{let $z_k = \left\{ 
              \begin{aligned}
                x_k - \hat{\mu}_+\quad & \text{if $k\in C_-$}\\
                \hat{\mu}_- - x_k\quad & \text{if $k\in C_+$}
              \end{aligned}
                                  \right. ,\, \forall k \in \{1,\cdots,N\}$}\\
  \multicolumn{2}{|l|}{let $A=\sum_{i=1}^N z_i z_i^T$}\\
  \multicolumn{2}{|l|}{let $v_1,\cdots,v_K$ be the eigenvectors of $A$} \\
  \multicolumn{2}{|l|}{$\quad$ with the largest eigenvalues}\\
  \multicolumn{2}{|l|}{let $P = [v_1,\cdots,v_K]$}\\
  \textbf{output}: & $\tilde{X} = XP$, $\tilde{Y} = Y$\\
\hline
\end{tabular}
\caption{The M-PCA1a(M-PCA1b) Algorithm}\label{figure-mpca1}
\end{center}
\end{figure}

\subsection{Algorithm: M-PCA2}

Our final algorithm, M-PCA2, attempts to capture the margin distribution more closely by simulating equation (\ref{eq-margin}) more closely. In this algorithm, we do not use all the pairs of data points in the two classes. Instead, we construct the difference vector between a datapoint and its nearest neighbor in the other class as a proxy for the margin distribution.

\begin{figure}[!h]
\begin{center}
\begin{tabular}{|ll|}
\hline
  \multicolumn{2}{|c|}{\textbf{M-PCA2}}\\
  \textbf{input}:& $X=(x_1,\cdots,x_N)^T$ \\
   & {$Y=(y_1,\cdots,y_N)^T$}\\
   & {target dimension $K$}\\
  \multicolumn{2}{|l|}{let $C_+=\{i:y_i=+1\},\, C_-=\{i:y_i=-1\}$}\\
  \multicolumn{2}{|l|}{let $z_{ij} = x_i - x_j, \forall i \in C_+, j \in C_-$, such that}\\
  \multicolumn{2}{|l|}{$\quad$ $i \in \argmin_{k\in C_+}\|x_k - x_j\|$ or $j \in \argmin_{k\in C_-}\|x_i-x_k\|$}\\
  \multicolumn{2}{|l|}{let $A=\sum_{i,j} z_{ij} z_{ij}^T$}\\
  \multicolumn{2}{|l|}{let $v_1,\cdots,v_K$ be the eigenvectors of $A$} \\
  \multicolumn{2}{|l|}{$\quad$ with the largest eigenvalues}\\
  \multicolumn{2}{|l|}{let $P = [v_1,\cdots,v_K]$}\\
  \textbf{output}: & $\tilde{X} = XP$, $\tilde{Y} = Y$\\
\hline
\end{tabular}
\caption{The M-PCA2 Algorithm}\label{figure-mpca2}
\end{center}
\end{figure}

Our experimental results show that M-PCA2 works extremely well, especially for small size high dimensional datasets. It is not hard to see that constructing the $\{z_{ij}\}$ takes $O(N^2D)$ operations and the size of $\{z_{ij}\}$ is no more than $N$. Therefore, the time complexity of M-PCA2 is the same as the vanilla PCA.

\section{Theory}\label{sec-theory}

Analyzing our schemes in full generality is difficult. To give some insight into how these work, in this section, we analyze a toy setting to provide some intuition, which suggests why our schemes can work better than PCA.

We consider a simple case where the classes have a shared class-conditional covariance matrix and for convenience we assume the class-conditional distributions are multivariate Gaussian, and the difference between the class means is aligned with one eigenvector of the class-conditional covariance matrix. Without loss of generality, we assume the class-conditional covariance matrix is diagonal. Let $X$ and $Y$ be two random variables such that
\begin{align}
\Pr(Y = +1) = \pi_+,\quad \Pr(Y=-1) = \pi_- = 1 - \pi_+,\\
X_+ = X | (Y=+1) \sim N(\mu_+, \Lambda),\\
X_- = X | (Y=-1) \sim N(\mu_-, \Lambda),
\end{align}
where
\begin{align}
0 < \pi_+ < 1,\\
\mu_+ = -\mu_- = (a,0,\cdots,0)^T,\quad a > 0,\\
\Lambda = \diag \{\lambda_1,\cdots,\lambda_D\}
%\Lambda = \begin{bmatrix} \lambda_1 & & \\
%                            & \ddots & \\
%                            & & \lambda_D
%          \end{bmatrix}
\end{align}
and where $\lambda_i,\, i=1,\cdots,D$ are not in any particular order. As a result, we have
\begin{equation}
X \sim N[\pi_+\mu_+ + \pi_-\mu_-,\, \Lambda + \pi_+\pi_-(\mu_+-\mu_-)(\mu_+-\mu_-)^T].
\end{equation}

It is clear in this setting that the only feature that is useful for classification is the first coordinate. A successful dimensionality reduction approach for this classification problem would have to include the first coordinate in the selected features. In the case $\pi_+ = \pi_- = \frac{1}{2}$, the optimal separating hyperplane is $x_1 = 0$ and the corresponding margin distribution is $N(a,\lambda_1)$. The whole margin distribution would be preserved if an approach includes the first coordinate in the reduced features.

Now suppose we sample $N$ data points $\{(x_i,y_i)\}_{i=1}^N$, then PCA works by eigen-decomposing
\begin{equation}
\frac{1}{N}\sum_{i=1}^N (x_i - \bar{x})(x_i - \bar{x})^T,
\end{equation}
where $\bar{x}$ is the sample mean. In expectation the above expression is just the covariance of the variable $X$, which is equal to
\begin{equation}
\diag \{ \lambda_1 + 4\pi_+\pi_- a^2, \lambda_2,\cdots,\lambda_D \}.
%\begin{bmatrix} \lambda_1 + 4\pi_+\pi_- a^2 & & & \\
%                                                               & \lambda_2 &  & \\
%                                                                 &  & \ddots & \\
%                                                              & & & \lambda_D
%                                \end{bmatrix}
\end{equation}
Suppose the target dimension is $K=1$, then for PCA to be successful, $\lambda_1 + 4\pi_+\pi_- a^2$ has to be larger than $\lambda_i,\, i=2,\cdots,D$. In most classification problems the between-class variance is usually larger than within-class variances, that is $a^2$ is usually large enough that $\lambda_1 + 4\pi_+\pi_- a^2$ is larger than $\lambda_i,\, i=2,\cdots,D$. This is why PCA can work well for classification tasks even though it is not designed for that purpose. In other situations, PCA may not work well. In particular here it will either be as good as possible or as bad as possible.

To see how M-PCA0 works, we define the random variable
\begin{equation}
Z^{(0)} := X_+ - X_- \sim N(\mu_+-\mu_-,\, 2\Lambda).
\end{equation}
M-PCA0 works by eigen-decomposing $\frac{1}{|C_+| |C_-|}\sum_{i \in C_-, j \in C_+} z_{ij} z_{ij}^T$. In expectation, this matrix is equal to
\begin{align}
E\left[ Z^{(0)} Z^{(0)T} \right] &= 2\Lambda + (\mu_+-\mu_-)(\mu_+-\mu_-)^T \nonumber\\
&= 2 \diag\{\lambda_1 + 2 a^2, \lambda_2,\cdots,\lambda_D \}
%2\begin{bmatrix} \lambda_1 + 2 a^2 & & & \\
%                                                               & \lambda_2 &  & \\
%                                                                 &  & \ddots & \\
%                                                              & & & \lambda_D
%                                \end{bmatrix}.
\end{align}
Since $\pi_+\pi_- \le \frac{1}{4}$, M-PCA0 has a better chance of selecting the first coordinate in the case $K=1$. Our experiments indeed shows that M-PCA0 usually performs better than PCA.

To see how M-PCA1a and M-PCA1b work, we use similar arguments. First, define the following random variable
\begin{equation}
Z^{(1)} := \left\{ 
              \begin{aligned}
                X_+ - \mu_- \quad & \text{with probability $\pi_+$},\\
                \mu_+ - X_- \quad & \text{with probability $\pi_-$}.
              \end{aligned}
                                  \right.
\end{equation}
From the definition, we have
\begin{equation}
Z^{(1)} \sim N(\mu_+-\mu_-,\, \Lambda).
\end{equation}
M-PCA1a and M-PCA1b work by eigen-decomposing $\frac{1}{N} \sum_{i=1}^N z_i z_i^T$, the expectation of which is equal to
\begin{align}
E\left[ Z^{(1)} Z^{(1)T} \right] &= \Lambda + (\mu_+-\mu_-)(\mu_+-\mu_-)^T \nonumber\\
&= \diag\{\lambda_1 + 4 a^2, \lambda_2,\cdots,\lambda_D \}
% \begin{bmatrix} \lambda_1 + 4 a^2 & & & \\
%                                                               & \lambda_2 &  & \\
%                                                                 &  & \ddots & \\
%                                                              & & & \lambda_D
%                                \end{bmatrix}.
\end{align}
This indicates that M-PCA1a and M-PCA1b have a much better chance than PCA of selecting the first coordinate as the useful feature in the case the target dimension $K$ is 1, since a larger quantity $4a^2$ is added to $\lambda_1$ rather than $4\pi_+\pi_- a^2 \le a^2$.

\begin{remark}
From the insight obtained above, PCA works poorly in the case of class imbalance due to the fact that the more class imbalance the smaller $4\pi_+\pi_- a^2$ becomes. Our schemes do not suffer from the same problem.
\end{remark}

\begin{remark}
Our schemes are much less vulnerable to the low between-class variance to within-class variance ratio problem that makes PCA fail. However, our schemes can still fail in the extreme cases. To make it even less vulnerable, we devise the scheme M-PCA2. M-PCA2 is based on our intuitive arguments.
\end{remark}

The situation quickly becomes intractable to analysis in the more general case that the difference between the class means is not aligned with any eigenvector of the class-conditional covariance matrix. In this case $\mu_+ = -\mu_- = (a_1,\cdots,a_D)$ with other quantities assumed the same as above. Now the covariance matrices to be eigen-decomposed are $\Lambda+\pi_+\pi_-(\mu_+-\mu_-)(\mu_+-\mu_-)^T$, $2\Lambda+(\mu_+-\mu_-)(\mu_+-\mu_-)^T$, $\Lambda+(\mu_+-\mu_-)(\mu_+-\mu_-)^T$ respectively. These are not diagonal and there is no simple analytical form for the eigenvectors. The most discriminative feature in theory is now FLD along the direction
\begin{equation}
\Lambda^{-1}(\mu_+ - \mu_-) = 2 \left(\frac{a_1}{\lambda_1}, \cdots, \frac{a_D}{\lambda_D}\right)^T
%\begin{bmatrix} a_1/\lambda_1 \\ \vdots \\ a_D/\lambda_D \end{bmatrix},
\end{equation}
which is not an eigenvector of the above matrices. We shall see from our experiments in Section \ref{sec-exp} that for small target dimension $K$, our approach nevertheless outperforms PCA in general, and despite being a (supervised) filter method, is highly competitive with wrapper approaches such as PLS and Lasso. %What is more, in real classification problems, the estimation of the covariance matrices is usually not precise, especially in the case of a small dataset size.

\section{Experiments}\label{sec-exp}

In this section, we present empirical results on the performance of our schemes: M-PCA0, M-PCA1a, M-PCA1b, and M-PCA2. For comparison, we compare them to vanilla PCA, PLS, and Lasso in terms of test errors. To obtain a comprehensive picture of the performance of the new schemes, we run experiments evaluating several widely used classifiers on a range of publicly available datasets with different characteristics. The classifiers used were Fisher Linear Discriminant (FLD), SVM, Logistic Regression (LR), and Naive Bayes (NB). We use two groups of publicly available real datasets. The first group contains datasets with the numbers of data points larger than the dimensions; these are \textit{ionosphere} \cite{Lichman:2013}, \textit{sonar} \cite{Lichman:2013}, \textit{mushrooms} \cite{Lichman:2013}, and \textit{splice} \cite{Lichman:2013}. The other group contains several small sample datasets with the numbers of data points smaller than the dimensions. These datasets are \textit{colon} \cite{Alon1999}, \textit{prostate} \cite{Singh2002}, \textit{ovarian} \cite{Schummer1999}, \textit{leukemia} \cite{Golub1999}, \textit{leukemia large} \cite{Golub1999}, and \textit{duke} \cite{West2001}. The information on these datasets is shown in Table \ref{table_datasets}.
\begin{table}[h!]
 \caption{Datasets}\label{table_datasets}
 \begin{center}
 \begin{tabular}{lccc}
  \toprule
   name            & source &  \#instances   &  \#features \\
  \midrule
   ionosphere      & \cite{Lichman:2013} &       126+225         &      34         \\
   sonar           & \cite{Lichman:2013} &       111+97          &      60         \\
   mushrooms       & \cite{Lichman:2013} &       3916+4208       &      112        \\
   splice          & \cite{Lichman:2013} &       1527+1648       &      60         \\
  \midrule
   colon           & \cite{Alon1999}     &       22+40           &      2000       \\
   prostate        & \cite{Singh2002}    &       50+52           &      6033       \\
   ovarian         & \cite{Schummer1999} &       24+30           &      1536       \\
   leukemia        & \cite{Golub1999}    &       47+25           &      3571       \\
   leukemia large  & \cite{Golub1999}    &       47+25           &      7129       \\
   duke            & \cite{West2001}     &       21+23           &      7129       \\
  \bottomrule
 \end{tabular}
 \end{center}
\end{table}

\subsection{Experiment Setup}

In the experiments, the test errors of combinations of a dimensionality reduction scheme, a target dimension $K$, a classifier and a dataset are obtained. Each combination of a dimensionality reduction scheme, a target dimension $K$, a classifier and a dataset is fed 50 independent partitions of the dataset into a training set and a test set, where four fifths of the data was used for training and the remainder for testing, and the sampling was stratified to preserve class membership proportions. Hence 50 test errors are produced for each combination. These are then used to compute the mean and the standard deviation of the test errors for that combination. For each loop iteration for a particular dataset, the data splits were held constant. We did not use cross validation since sign test assumes independent observations. %The combinations with a same dataset is fed with the same set of random partitions of the dataset. The random partitions are obtained such that $\frac{4}{5}$ of the dataset is used for training and the remaining used for testing, and that the proportions of the datapoints belonging to each class are preserved.

We choose two representative target dimensions for each dataset. For small sample $(N<D)$ datasets, the target dimensions are $\frac{R}{4}$ and $\frac{R}{2}$, where $R$ is the rank of the training data matrix, which is roughly four fifths of the number of instances. While for the other datasets $(N>D)$, the target dimensions are chosen to be $\frac{D}{6}$ and $\frac{D}{3}$, where $D$ is the number of features (original data dimension). We do not include here larger target dimensions. This is because, on one hand, small target dimensions are the settings of practical interest. On the other hand, with increased target dimension $K$, the differences between different dimensionality reduction schemes become less obvious. This is due to that fact that as the target dimension $K$ approaches the rank of the training data matrix, the number $K$ of features retained by different schemes are large enough to incorporate almost all the discriminative information. Our own experiments, not presented here, also indicate this fact.

For the datasets with more features than instances, the test errors of M-PCA0, M-PCA1a, M-PCA1b, and M-PCA2 are compared with that of PCA, PLS, and Lasso. While for the datasets with fewer features than instances, due to the high computational cost of the M-PCA0 scheme, only M-PCA1a, M-PCA1b, and M-PCA2 are run and compared to PCA, PLS, and Lasso. We use the SVM and Logistic Regression implemented by \textit{liblinear} \cite{Fan2008}. The version of SVM classifier used for experiment is the $\ell_2$-regularized $\ell_2$-loss SVM. While the logistic regression classifier used here is $\ell_2$-regularized. The classifier parameters used were selected by cross validation on the whole original dataset to provide a consistent baseline, even though this may have favored the full wrapper methods since they have access to the tuned classifier. The Naive Bayes model uses a Gaussian class-conditional likelihood model and we used the built-in MATLAB version.

Finally, the way we run PLS and Lasso to reduce dimensionality for the classification tasks is by casting the classification tasks as regression tasks with discrete targets and obtain the important features. The implementations of PLS and Lasso used are the built-in MATLAB functions \textit{plsregress} and \textit{lasso}.

\subsection{Results}

\begin{figure}[t]
\begin{center}
\includegraphics[width=0.8\textwidth]{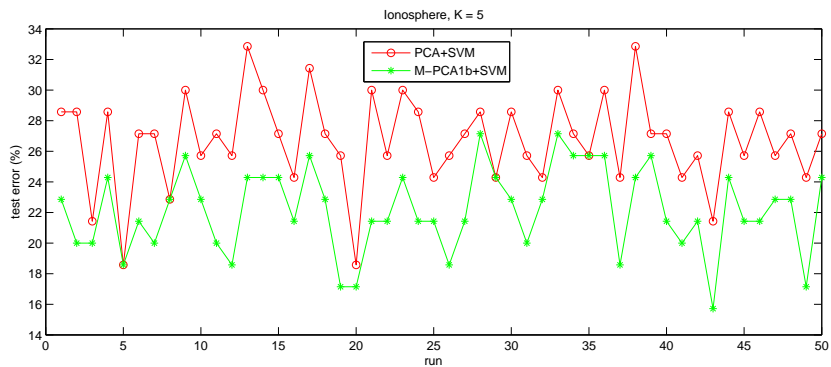}
\end{center}
\caption{Representative example plot of test errors for PCA-vs-M-PCA1b with SVM classifier for Ionosphere with K = 5.}\label{lineplot}
\end{figure}

\begin{table}[!ht]
%\captionsetup{width=0.45\textwidth,font=footnotesize}
\caption{Non-parametric sign test on per data split test errors for PCA vs. our variants on $N>D$ datasets. All $p$-values correspond to $H_0$: PCA test error is strictly smaller than our variants. The boldfaced numbers indicate the cases where $p<0.05$.} \label{pvalue-large}
\begin{center}
\scalebox{0.8}{
 \begin{tabular}{ccccc}
  \toprule
   K & Classifier & M-PCA1a & M-PCA1b & M-PCA2 \\
  \midrule
   \multicolumn{5}{c}{Ionosphere} \\
  \midrule
  \multirow{2}{*}{5}   & SVM &  0.3318 &  $\mathbf{1.4211\times10^{-14}}$ &  0.6864 \\* %\cline{2-5}
                       & LR  & 0.8569  &  $\mathbf{2.1778\times10^{-12}}$ &  0.9878 \\*
  \midrule
  \multirow{2}{*}{11}  & SVM &  $\mathbf{0.0096}$ &  $\mathbf{2.8422\times10^{-14}}$ &  $\mathbf{1.4144\times10^{-6}}$ \\* %\cline{2-5}
                       & LR  &  $\mathbf{0.0154}$ &  $\mathbf{6.5629\times10^{-11}}$ &  $\mathbf{3.7645\times10^{-4}}$ \\*
  \midrule
   \multicolumn{5}{c}{Sonar} \\
  \midrule
  \multirow{2}{*}{10}  & SVM & 0.5  & 0.9981  &  0.6911 \\* %\cline{2-5}
                       & LR  &  0.2272 &  0.9061 &  0.9061 \\*
  \midrule
  \multirow{2}{*}{20}  & SVM &  0.998 &  0.1215 &  0.111 \\* %\cline{2-5}
                       & LR  &  0.9988 &  0.6358 & 0.2664  \\*
  \midrule
   \multicolumn{5}{c}{Mushrooms} \\
  \midrule
  \multirow{2}{*}{18}  & SVM & $\mathbf{1.7764\times10^{-15}}$  &  0.7336 &  $\mathbf{8.8818\times10^{-16}}$ \\* %\cline{2-5}
                       & LR  & $\mathbf{8.8818\times10^{-16}}$  & 0.1856  &  $\mathbf{8.8818\times10^{-16}}$ \\*
  \midrule
  \multirow{2}{*}{37}  & SVM &  $\mathbf{2.558\times10^{-12}}$ & $\mathbf{8.8818\times10^{-16}}$  &  $\mathbf{8.8818\times10^{-16}}$ \\* %\cline{2-5}
                       & LR  & $\mathbf{4.5475\times10^{-13}}$  &  $\mathbf{0.004}$ & $\mathbf{8.8818\times10^{-16}}$  \\*
  \midrule
   \multicolumn{5}{c}{Splice} \\
  \midrule
  \multirow{2}{*}{10}  & SVM & $\mathbf{8.8818\times10^{-14}}$  & 1  &  $\mathbf{9.8233\times10^{-7}}$ \\* %\cline{2-5}
                       & LR  &  $\mathbf{1.1333\times10^{-12}}$ & 1  &  $\mathbf{2.5501\times10^{-9}}$ \\*
  \midrule
  \multirow{2}{*}{20}  & SVM & $\mathbf{7.5692\times10^{-10}}$  & 1  &  $\mathbf{3.0848\times10^{-5}}$ \\* %\cline{2-5}
                       & LR  &  $\mathbf{8.8818\times10^{-14}}$ &  1 &  $\mathbf{1.73\times10^{-4}}$ \\*
  \bottomrule
 \end{tabular}
}
\end{center}
\end{table}

\begin{table}[!ht]
%\captionsetup{width=0.45\textwidth,font=footnotesize}
\caption{Non-parametric sign test on per data split test errors for PCA vs. our variants on $N<D$ datasets. All $p$-values correspond to $H_0$: PCA test error is strictly smaller than our variants. The boldfaced numbers indicate the cases where $p<0.05$.} \label{pvalue-small}
\begin{center}
\scalebox{0.8}{
 \begin{tabular}{cccccc}
  \toprule
   K & Classifier & M-PCA0 & M-PCA1a & M-PCA1b & M-PCA2\\
  \midrule
   \multicolumn{6}{c}{Colon} \\
  \midrule
  \multirow{2}{*}{12}  & SVM & 0.25  &  0.75 & 0.25 & 0.3125 \\* %\cline{2-6}
                       & LR  & 0.9648  & 0.7734  & 0.1445 & 0.3036 \\*
  \midrule
  \multirow{2}{*}{24}  & SVM &  0.5 &  0.125 & 0.5 & $\mathbf{0.0078}$ \\* %\cline{2-6}
                       & LR  & $\mathbf{2.4414\times10^{-4}}$  &  $\mathbf{4.8828\times10^{-4}}$ & $\mathbf{0.0017}$ & 0.1051 \\*
  \midrule
   \multicolumn{6}{c}{Prostate} \\
  \midrule
  \multirow{2}{*}{20}  & SVM & 0.5  &  0.212 & 0.0898 & $\mathbf{0.0145}$ \\* %\cline{2-6}
                       & LR  & 0.0898  &  $\mathbf{0.0106}$ & $\mathbf{0.0036}$ & $\mathbf{1.372\times10^{-5}}$ \\*
  \midrule
  \multirow{2}{*}{40}  & SVM & 0.5  & 0.2266  & 0.3633 & 0.2272 \\* %\cline{2-6}
                       & LR  & 0.5  &  0.5 & 0.5 & 0.1509 \\*
  \midrule
   \multicolumn{6}{c}{Ovarian} \\
  \midrule
  \multirow{2}{*}{10}  & SVM &  0.6762 &  0.6762 & 0.968 & 0.1635 \\* %\cline{2-6}
                       & LR  & 0.2905  &  $\mathbf{0.0021}$ & 0.1316 & $\mathbf{1.5162\times10^{-6}}$ \\*
  \midrule
  \multirow{2}{*}{21}  & SVM & 0.8867  &  0.8867 & 0.9961 & 0.2272 \\* %\cline{2-6}
                       & LR  &  0.8555 & 0.8555  & 0.9648 & $\mathbf{0.0053}$ \\*
  \midrule
   \multicolumn{6}{c}{Leukemia} \\
  \midrule
  \multirow{2}{*}{14}  & SVM & 0.5  & 0.9375  & 0.875 & 0.2539 \\* %\cline{2-6}
                       & LR  &  0.3125 &  0.5 & 0.1094 & $\mathbf{0.002}$ \\*
  \midrule
  \multirow{2}{*}{28}  & SVM &  1 &  1 & 0.5 & 0.9375 \\*  %\cline{2-6}
                       & LR  &  0.5 & 0.5  & 0.875 & 0.6875 \\*
  \midrule
   \multicolumn{6}{c}{Leuk-large} \\
  \midrule
  \multirow{2}{*}{14}  & SVM & 0.1133  & 0.0547  & 0.1719 & 0.5982 \\* %\cline{2-6}
                       & LR  & 0.8867  &  0.1334 & 0.1938 & $\mathbf{0.0245}$ \\*
  \midrule
  \multirow{2}{*}{28}  & SVM &  0.1875 &  0.5 & 0.5 & 0.3633 \\* %\cline{2-6}
                       & LR  & 0.0625  & $\mathbf{0.0312}$ & $\mathbf{0.0078}$ & 0.5 \\* 
  \midrule
   \multicolumn{6}{c}{Duke} \\
  \midrule
  \multirow{2}{*}{8}   & SVM &  0.073 & 0.1509  & 0.1334 & 0.3318 \\* %\cline{2-6}
                       & LR  &  $\mathbf{0.0064}$ & $\mathbf{0.0064}$  & $\mathbf{0.0038}$ & 0.068 \\*
  \midrule
  \multirow{2}{*}{17}  & SVM & 0.125  & 0.125  & 0.3125 & 0.3872 \\* %\cline{2-6}
                       & LR  & 0.1875  & 0.3438  & 0.2266 & 0.337 \\*
  \bottomrule
 \end{tabular}
}
\end{center}
\end{table}

The detailed results are in the appendix and shown in Table~\ref{table-iono} - \ref{table-duke}. As can be seen from the tables, our schemes are superior to PCA in general and highly competitive with PLS and Lasso. This indicates that our schemes are indeed able to select the most discriminative features. Due to the large variance of the test errors\footnote{This is more obvious for small sample datasets due to smaller number of observations and the fact that the test error is very sensitive to different partitions of the original datasets.}, it appears initially that our schemes are not statistically significantly better than PCA, however these averages are across the different partitions of the datasets and performance on the different individual data splits is highly variable. To see if our approach indeed significantly outperforms PCA, we compare test errors on independent data splits and use a non-parametric sign test to evaluate the null: See Figure~\ref{lineplot} -- Each test error corresponds to an independent data split. We see that for this dataset M-PCA1b is at least as good as PCA on every individual split, but the high between-sample variance in the test error will mask this fact if we only consider the average test error across data splits. We tabulate the sign test p-values of our algorithms versus PCA in Tables~\ref{pvalue-large} and \ref{pvalue-small}. Here we only show results on SVM and Logistic Regression. The boldfaced numbers indicate the cases where $p<0.05$ and we have made no correction to these values for multiple comparisons since the smallest p-values are generally below $0.0125$ in any case. We see that our approach does significantly outperform PCA, especially when the number of retained dimensions is small.

\clearpage
\section{Discussion}

We presented four simple filter PCA variants, each of which seems to improve the performance on projected data in a classification task over standard PCA. Our ideas in this paper are primarily based on the observation that the margin distribution contains most of the discriminative information, for a classification task and the PCA objective is not obviously aligned with preserving this quantity. Therefore we propose four heuristic structures to represent the margin distribution on which we then perform PCA. Extensive empirical evaluations suggest these do indeed represent the margin distribution well. Whether there are some better structures for representing the margin distribution that can be evaluated outside of a wrapper approach is an interesting question for future research. Any such better structures should give us a better dimensionality reduction scheme for linear classification. Whether theoretical guarantees on classification performance using our approach are possible looks like a difficult open problem - the main hurdle is how to provide typical case guarantees for a deterministic algorithm without imposing restrictive conditions on the data generator. A further direction for future research is to consider non-linear dimensionality reduction schemes, such as kernel PCA in a similar light.

\appendix
\section{Detailed Experimental Results}

In the tables below, the test errors (mean$\pm$std) are shown in percent and the boldfaced numbers represent the best performing schemes.
\begin{table}[!ht]
%\captionsetup{width=0.5\textwidth,font=footnotesize}
\caption{Ionosphere Test Errors} \label{table-iono}
\begin{center}
\scalebox{0.9}{
 \begin{tabular}{clcccc}
  \toprule
   K & Scheme & SVM & LR & FLD & NB \\
  \midrule
  \multirow{6}{*}{5}   & PCA     & 26.7$\pm$3.0 & 26.4$\pm$2.8 & 17.0$\pm$3.7 & $\mathbf{12.3\pm3.8}$ \\*
                       & PLS     & 23.5$\pm$3.0 & 23.4$\pm$3.2 & $\mathbf{11.5\pm3.0}$ & 20.7$\pm$5.1 \\*
                       & Lasso   & 27.7$\pm$2.6 & 27.0$\pm$2.8 & 20.0$\pm$3.9 & 13.2$\pm$3.7 \\*
                       & M-PCA1a & 26.6$\pm$3.0 & 26.5$\pm$2.8 & 16.4$\pm$3.8 & 17.2$\pm$4.3 \\*
                       & M-PCA1b & $\mathbf{22.1\pm2.8}$ & $\mathbf{21.6\pm3.5}$ & 16.7$\pm$3.6 & 17.5$\pm$4.3 \\*
                       & M-PCA2  & 26.9$\pm$2.4 & 27.1$\pm$2.6 & 21.3$\pm$4.0 & 18.8$\pm$4.7 \\*
  \midrule
  \multirow{6}{*}{11}  & PCA     & 25.3$\pm$2.6 & 24.4$\pm$2.7 & 18.4$\pm$3.6 & 10.4$\pm$3.7 \\*
                       & PLS     & 20.1$\pm$3.7 & 19.5$\pm$3.3 & $\mathbf{13.2\pm3.1}$ & 14.8$\pm$4.3 \\*
                       & Lasso   & $\mathbf{19.8\pm4.0}$ & $\mathbf{19.1\pm4.2}$ & 20.2$\pm$3.4 & $\mathbf{10.1\pm3.7}$ \\*
                       & M-PCA1a & 24.7$\pm$2.9 & 24.1$\pm$2.7 & 18.5$\pm$3.7 & 11.2$\pm$3.9 \\*
                       & M-PCA1b & 21.1$\pm$2.9 & 20.4$\pm$3.2 & 18.0$\pm$3.8 & 11.1$\pm$3.9 \\*
                       & M-PCA2  & 23.5$\pm$3.2 & 23.2$\pm$2.9 & 20.7$\pm$3.6 & 15.2$\pm$5.0 \\*
  \bottomrule
 \end{tabular}
}
\end{center}
\end{table}
\begin{table}[!ht]
%\captionsetup{width=0.5\textwidth,font=footnotesize}
\caption{Sonar Test Errors} \label{table-sonar}
\begin{center}
\scalebox{0.9}{
 \begin{tabular}{clcccc}
  \toprule
   K & Scheme & SVM & LR & FLD & NB \\
  \midrule
  \multirow{6}{*}{10}  & PCA     & 26.8$\pm$5.7 & 25.9$\pm$6.2 & 24.4$\pm$6.7 & $\mathbf{22.6\pm6.4}$ \\*
                       & PLS     & 30.2$\pm$7.0 & 29.0$\pm$6.5 & 25.6$\pm$6.8 & 37.5$\pm$8.1 \\*
                       & Lasso   & 30.9$\pm$5.9 & 27.8$\pm$6.3 & 24.6$\pm$6.2 & 32.8$\pm$6.5 \\*
                       & M-PCA1a & $\mathbf{26.7\pm5.6}$ & $\mathbf{25.4\pm6.1}$ & $\mathbf{23.6\pm6.6}$ & 24.3$\pm$6.5 \\*
                       & M-PCA1b & 27.8$\pm$5.5 & 26.5$\pm$6.1 & 24.4$\pm$6.6 & 24.9$\pm$6.3 \\*
                       & M-PCA2  & 27.6$\pm$5.7 & 26.1$\pm$6.0 & 23.8$\pm$6.6 & 24.8$\pm$6.9 \\*
  \midrule
  \multirow{6}{*}{20}  & PCA     & 27.5$\pm$5.7 & 25.8$\pm$6.7 & $\mathbf{23.6\pm6.6}$ & $\mathbf{23.5\pm6.9}$ \\*
                       & PLS     & 27.6$\pm$6.2 & 27.9$\pm$5.4 & 28.2$\pm$6.1 & 38.5$\pm$8.4 \\*
                       & Lasso   & 30.1$\pm$5.5 & 29.3$\pm$7.0 & 27.1$\pm$5.4 & 32.1$\pm$7.2 \\*
                       & M-PCA1a & 28.3$\pm$5.7 & 26.7$\pm$6.2 & 23.8$\pm$7.0 & 26.0$\pm$6.0 \\*
                       & M-PCA1b & 27.1$\pm$5.9 & 26.0$\pm$6.2 & 23.7$\pm$6.7 & 28.2$\pm$8.0 \\*
                       & M-PCA2  & $\mathbf{26.7\pm5.7}$ & $\mathbf{25.2\pm6.4}$ & 24.0$\pm$6.1 & 27.2$\pm$8.2 \\*
  \bottomrule
 \end{tabular}
}
\end{center}
\end{table}
\begin{table}[!ht]
%\captionsetup{width=0.5\textwidth,font=footnotesize}
\caption{Mushrooms Test Errors} \label{table-mushrooms}
\begin{center}
\scalebox{0.9}{
 \begin{tabular}{clcccc}
  \toprule
   K & Scheme & SVM & LR & FLD & NB \\
  \midrule
  \multirow{6}{*}{18}  & PCA     & 3.2$\pm$0.3 & 3.1$\pm$0.4 & 3.8$\pm$0.5 & 11.4$\pm$0.5 \\*
                       & PLS     & 0.5$\pm$0.3 & 0.7$\pm$0.4 & $\mathbf{0.2\pm0.1}$ & 11.9$\pm$0.7 \\*
                       & Lasso   & $\mathbf{0.1\pm0.1}$ & $\mathbf{0.2\pm0.2}$ & 1.6$\pm$1.1 & $\mathbf{6.2\pm1.2}$ \\*
                       & M-PCA1a & 2.8$\pm$0.4 & 2.7$\pm$0.4 & 3.5$\pm$0.5 & 11.7$\pm$0.6 \\*
                       & M-PCA1b & 3.2$\pm$0.3 & 3.0$\pm$0.4 & 3.1$\pm$0.4 & 11.4$\pm$0.6 \\*
                       & M-PCA2  & 1.5$\pm$0.4 & 1.7$\pm$0.3 & 2.7$\pm$0.4 & 10.7$\pm$0.9 \\*
  \midrule
  \multirow{6}{*}{37}  & PCA     & 2.0$\pm$0.3 & 2.0$\pm$0.3 & 3.0$\pm$0.5 & 10.0$\pm$0.7 \\*
                       & PLS     & 0.2$\pm$0.1 & 0.7$\pm$0.6 & $\mathbf{0.1\pm0.2}$ & 12.1$\pm$0.8 \\*
                       & Lasso   & $\mathbf{0.1\pm0.1}$ & $\mathbf{0.1\pm0.1}$ & 1.3$\pm$0.8 & $\mathbf{6.1\pm0.6}$ \\*
                       & M-PCA1a & 1.8$\pm$0.3 & 1.9$\pm$0.3 & 2.8$\pm$0.5 & 10.8$\pm$0.6 \\*
                       & M-PCA1b & 1.3$\pm$0.3 & 2.0$\pm$0.3 & 3.1$\pm$0.5 & 8.5$\pm$0.6 \\*
                       & M-PCA2  & 0.3$\pm$0.1 & 0.4$\pm$0.2 & 1.7$\pm$0.5 & 11.1$\pm$0.9 \\*
  \bottomrule
 \end{tabular}
}
\end{center}
\end{table}
\begin{table}[!ht]
%\captionsetup{width=0.5\textwidth,font=footnotesize}
\caption{Splice Test Errors} \label{table-splice}
\begin{center}
\scalebox{0.9}{
 \begin{tabular}{clcccc}
  \toprule
   K & Scheme & SVM & LR & FLD & NB \\
  \midrule
  \multirow{6}{*}{10}  & PCA     & 18.2$\pm$1.3 & 18.2$\pm$1.1 & 17.9$\pm$1.3 & 16.1$\pm$1.3 \\*
                       & PLS     & $\mathbf{16.1\pm1.3}$ & $\mathbf{16.2\pm1.2}$ & $\mathbf{15.4\pm1.2}$ & 22.4$\pm$1.5 \\*
                       & Lasso   & 18.4$\pm$1.3 & 18.5$\pm$1.9 & 18.0$\pm$1.6 & 15.1$\pm$1.6 \\*
                       & M-PCA1a & 16.5$\pm$1.3 & 16.4$\pm$1.2 & 16.0$\pm$1.2 & 18.1$\pm$1.5 \\*
                       & M-PCA1b & 22.0$\pm$1.4 & 22.3$\pm$1.6 & 21.4$\pm$1.8 & 22.2$\pm$1.7 \\*
                       & M-PCA2  & 17.0$\pm$1.3 & 16.8$\pm$1.3 & 16.2$\pm$1.2 & $\mathbf{14.4\pm1.3}$ \\*
  \midrule
  \multirow{6}{*}{20}  & PCA     & 17.9$\pm$1.3 & 17.7$\pm$1.0 & 17.1$\pm$1.5 & 16.0$\pm$1.5 \\*
                       & PLS     & $\mathbf{16.0\pm1.3}$ & $\mathbf{16.2\pm1.2}$ & 15.4$\pm$1.2 & 22.3$\pm$1.6 \\*
                       & Lasso   & 16.3$\pm$1.4 & 16.2$\pm$1.3 & $\mathbf{15.3\pm1.1}$ & $\mathbf{13.0\pm1.3}$ \\*
                       & M-PCA1a & 16.8$\pm$1.3 & 16.4$\pm$1.2 & 15.9$\pm$1.3 & 18.0$\pm$1.6 \\*
                       & M-PCA1b & 21.0$\pm$1.3 & 21.1$\pm$1.3 & 20.0$\pm$1.8 & 21.0$\pm$1.7 \\*
                       & M-PCA2  & 17.0$\pm$1.3 & 16.8$\pm$1.3 & 16.2$\pm$1.3 & 14.3$\pm$1.2 \\*
  \bottomrule
 \end{tabular}
}
\end{center}
\end{table}
\begin{table}[!ht]
%\captionsetup{width=0.5\textwidth,font=footnotesize}
\caption{Colon Test Errors} \label{table-colon}
\begin{center}
\scalebox{0.9}{
 \begin{tabular}{clcccc}
  \toprule
   K & Scheme & SVM & LR & FLD & NB \\
  \hline
  \multirow{7}{*}{12}  & PCA     & 11.7$\pm$9.1 & 13.5$\pm$8.9 & $\mathbf{12.0\pm7.9}$ & 17.5$\pm$8.6 \\*
                       & PLS     & 14.0$\pm$9.3 & 13.2$\pm$8.9 & 14.7$\pm$7.3 & 17.8$\pm$8.6 \\*
                       & Lasso   & 21.2$\pm$11.9 & 13.2$\pm$8.9 & 22.2$\pm$8.8 & $\mathbf{13.5\pm8.7}$ \\*
                       & M-PCA0  & 11.3$\pm$9.2 & 14.0$\pm$9.0 & 12.2$\pm$8.1 & 21.8$\pm$9.8 \\*
                       & M-PCA1a & 11.7$\pm$9.1 & 13.7$\pm$9.3 & 12.2$\pm$7.9 & 15.8$\pm$8.5 \\*
                       & M-PCA1b & $\mathbf{11.3\pm9.0}$ & $\mathbf{12.7\pm9.3}$ & 12.2$\pm$8.5 & 16.3$\pm$8.6 \\*
                       & M-PCA2  & 11.3$\pm$9.2 & 12.8$\pm$9.1 & 12.7$\pm$8.6 & 14.3$\pm$8.4 \\*
  \midrule
  \multirow{7}{*}{24}  & PCA     & 12.8$\pm$9.4 & 14.8$\pm$9.1 & 13.7$\pm$8.2 & 20.2$\pm$9.5 \\*
                       & PLS     & 15.3$\pm$9.7 & 14.5$\pm$9.8 & 14.0$\pm$7.6 & 21.5$\pm$8.6 \\*
                       & Lasso   & 23.5$\pm$11.0 & 15.2$\pm$8.7 & 22.2$\pm$8.8 & $\mathbf{13.7\pm7.7}$ \\*
                       & M-PCA0  & 12.5$\pm$9.1 & $\mathbf{12.8\pm8.6}$ & $\mathbf{13.0\pm8.1}$ & 20.0$\pm$8.7 \\*
                       & M-PCA1a & 12.3$\pm$9.4 & 13.0$\pm$9.8 & 13.3$\pm$8.1 & 15.7$\pm$8.4 \\*
                       & M-PCA1b & 12.7$\pm$9.6 & 13.0$\pm$9.4 & 14.0$\pm$7.6 & 15.5$\pm$8.4 \\*
                       & M-PCA2  & $\mathbf{11.3\pm8.5}$ & 13.8$\pm$8.8 & 13.8$\pm$7.3 & 14.0$\pm$8.5 \\*
  \bottomrule
 \end{tabular}
}
\end{center}
\end{table}
\begin{table}[!ht]
%\captionsetup{width=0.5\textwidth,font=footnotesize}
\caption{Prostate Test Errors} \label{table-prostate}
\begin{center}
\scalebox{0.9}{
 \begin{tabular}{clcccc}
  \toprule
   K & Scheme & SVM & LR & FLD & NB \\
  \midrule
  \multirow{7}{*}{20}  & PCA     & 10.1$\pm$5.2 & 10.4$\pm$6.9 & 9.0$\pm$4.8 & 33.9$\pm$12.8 \\*
                       & PLS     & 9.2$\pm$5.1 & 8.2$\pm$6.1 & 8.5$\pm$4.8 & 43.9$\pm$7.8 \\*
                       & Lasso   & 9.3$\pm$6.1 & 8.3$\pm$5.9 & 10.3$\pm$5.6 & $\mathbf{8.1\pm5.9}$ \\*
                       & M-PCA0  & 9.9$\pm$5.3 & 9.8$\pm$6.8 & 8.8$\pm$5.0 & 39.7$\pm$9.9 \\*
                       & M-PCA1a & 9.4$\pm$4.7 & 9.2$\pm$6.6 & 8.7$\pm$5.0 & 42.7$\pm$8.8 \\*
                       & M-PCA1b & 9.1$\pm$4.8 & 8.9$\pm$5.9 & 8.0$\pm$5.1 & 35.0$\pm$9.0 \\*
                       & M-PCA2  & $\mathbf{8.4\pm4.6}$ & $\mathbf{7.3\pm6.8}$ & $\mathbf{7.5\pm4.7}$ & 30.3$\pm$11.0 \\*
  \midrule
  \multirow{7}{*}{40}  & PCA     & 8.3$\pm$4.7 & 8.0$\pm$6.1 & 7.5$\pm$4.7 & 42.1$\pm$8.9 \\*
                       & PLS     & 9.3$\pm$4.8 & 8.4$\pm$6.1 & 8.5$\pm$4.8 & 46.5$\pm$5.4 \\*
                       & Lasso   & $\mathbf{7.7\pm5.2}$ & $\mathbf{6.2\pm5.0}$ & 10.0$\pm$5.4 & $\mathbf{9.5\pm5.6}$ \\*
                       & M-PCA0  & 8.2$\pm$4.8 & 7.9$\pm$6.0 & 7.2$\pm$4.6 & 45.0$\pm$6.7 \\*
                       & M-PCA1a & 8.0$\pm$4.7 & 7.9$\pm$6.0 & $\mathbf{7.2\pm4.5}$ & 47.3$\pm$5.4 \\*
                       & M-PCA1b & 8.1$\pm$4.9 & 7.9$\pm$5.5 & 7.4$\pm$4.9 & 42.7$\pm$7.8 \\*
                       & M-PCA2  & 7.9$\pm$5.2 & 7.4$\pm$6.4 & 7.2$\pm$5.2 & 33.4$\pm$11.3 \\*
  \bottomrule
 \end{tabular}
}
\end{center}
\end{table}
\begin{table}[!ht]
%\captionsetup{width=0.5\textwidth,font=footnotesize}
\caption{Ovarian Test Errors} \label{table-ovarian}
\begin{center}
\scalebox{0.9}{
 \begin{tabular}{clcccc}
  \toprule
   K & Scheme & SVM & LR & FLD & NB \\
  \midrule
  \multirow{7}{*}{10}  & PCA     & 18.5$\pm$10.9 & 23.1$\pm$10.6 & 26.0$\pm$11.6 & 45.8$\pm$10.4 \\*
                       & PLS     & 19.1$\pm$10.7 & 17.3$\pm$8.9 & $\mathbf{17.1\pm9.8}$ & 48.4$\pm$11.5 \\*
                       & Lasso   & 22.5$\pm$13.3 & 26.7$\pm$10.5 & 27.8$\pm$11.4 & $\mathbf{25.8\pm13.0}$ \\*
                       & M-PCA0  & 18.5$\pm$10.4 & 22.4$\pm$10.6 & 23.6$\pm$11.2 & 45.3$\pm$13.6 \\*
                       & M-PCA1a & 18.4$\pm$10.1 & 20.7$\pm$10.9 & 22.2$\pm$12.2 & 47.6$\pm$11.8 \\*
                       & M-PCA1b & 20.2$\pm$10.6 & 21.8$\pm$11.6 & 22.0$\pm$11.2 & 47.3$\pm$11.8 \\*
                       & M-PCA2  & $\mathbf{17.1\pm10.2}$ & $\mathbf{17.1\pm7.9}$ & 20.4$\pm$10.2 & 33.6$\pm$12.5 \\*
  \midrule
  \multirow{7}{*}{21}  & PCA     & 18.7$\pm$10.5 & 19.1$\pm$9.6 & 20.0$\pm$10.1 & 42.5$\pm$11.4 \\*
                       & PLS     & $\mathbf{17.1\pm10.5}$ & $\mathbf{15.5\pm9.4}$ & $\mathbf{15.1\pm10.2}$ & 50.4$\pm$8.3 \\*
                       & Lasso   & 18.5$\pm$11.9 & 24.5$\pm$11.7 & 24.9$\pm$13.3 & $\mathbf{27.5\pm13.8}$ \\*
                       & M-PCA0  & 19.3$\pm$10.3 & 19.5$\pm$9.2 & 19.1$\pm$10.9 & 44.2$\pm$11.8 \\*
                       & M-PCA1a & 19.3$\pm$11.1 & 19.5$\pm$9.7 & 18.9$\pm$11.1 & 45.1$\pm$11.6 \\*
                       & M-PCA1b & 19.8$\pm$11.0 & 19.8$\pm$9.3 & 21.3$\pm$11.4 & 45.8$\pm$11.3 \\*
                       & M-PCA2  & 18.0$\pm$10.1 & 16.0$\pm$9.5 & 19.6$\pm$12.0 & 29.3$\pm$12.4 \\*
  \bottomrule
 \end{tabular}
}
\end{center}
\end{table}
\begin{table}[!ht]
%\captionsetup{width=0.5\textwidth,font=footnotesize}
\caption{Leukemia Test Errors} \label{table-leukemia}
\begin{center}
\scalebox{0.9}{
 \begin{tabular}{clcccc}
  \toprule
   K & Scheme & SVM & LR & FLD & NB \\
  \midrule
  \multirow{7}{*}{14}  & PCA     & 3.1$\pm$3.6 & 3.4$\pm$3.9 & 2.0$\pm$3.2 & 2.4$\pm$3.7 \\*
                       & PLS     & 4.0$\pm$4.1 & 3.1$\pm$4.1 & 2.9$\pm$3.5 & 3.4$\pm$4.8 \\*
                       & Lasso   & 5.9$\pm$4.9 & 4.1$\pm$5.0 & 5.0$\pm$6.2 & 4.4$\pm$5.2 \\*
                       & M-PCA0  & 3.0$\pm$3.6 & 3.1$\pm$4.1 & $\mathbf{1.9\pm3.2}$ & $\mathbf{2.0\pm3.2}$ \\*
                       & M-PCA1a & 3.4$\pm$3.6 & 3.3$\pm$4.1 & 2.0$\pm$3.2 & 2.4$\pm$3.4 \\*
                       & M-PCA1b & 3.3$\pm$3.6 & 2.9$\pm$4.1 & 2.0$\pm$3.2 & 2.6$\pm$3.5 \\*
                       & M-PCA2  & $\mathbf{2.7\pm3.5}$ & $\mathbf{2.1\pm3.6}$ & 2.1$\pm$3.3 & 2.7$\pm$4.1 \\*
  \midrule
  \multirow{7}{*}{28}  & PCA     & 2.4$\pm$3.4 & 2.4$\pm$4.0 & 2.0$\pm$3.2 & 2.9$\pm$4.6 \\*
                       & PLS     & 4.7$\pm$4.2 & 3.3$\pm$4.4 & 2.9$\pm$3.5 & 3.6$\pm$4.8 \\*
                       & Lasso   & 6.0$\pm$5.3 & 4.6$\pm$4.9 & 6.0$\pm$6.7 & 4.3$\pm$4.8 \\*
                       & M-PCA0  & 2.9$\pm$3.5 & $\mathbf{2.3\pm3.7}$ & 1.9$\pm$3.2 & 4.7$\pm$5.7 \\*
                       & M-PCA1a & 2.4$\pm$3.4 & 2.3$\pm$3.9 & 2.0$\pm$3.2 & $\mathbf{2.1\pm3.3}$ \\*
                       & M-PCA1b & $\mathbf{2.3\pm3.4}$ & 2.6$\pm$4.0 & $\mathbf{1.9\pm3.2}$ & 2.3$\pm$3.4 \\*
                       & M-PCA2  & 2.7$\pm$3.5 & 2.4$\pm$4.0 & 2.1$\pm$3.3 & 3.0$\pm$3.8 \\*
  \bottomrule
 \end{tabular}
}
\end{center}
\end{table}
\begin{table}[!ht]
%\captionsetup{width=0.5\textwidth,font=footnotesize}
\caption{Leuk-large Test Errors} \label{table-leuklarge}
\begin{center}
\scalebox{0.9}{
 \begin{tabular}{clcccc}
  \toprule
   K & Scheme & SVM & LR & FLD & NB \\
  \midrule
  \multirow{7}{*}{14}  & PCA     & 5.7$\pm$5.6 & 6.1$\pm$4.8 & 3.9$\pm$4.4 & 11.0$\pm$7.2 \\*
                       & PLS     & 4.7$\pm$4.5 & 5.0$\pm$5.4 & 3.9$\pm$5.6 & 32.6$\pm$13.6 \\*
                       & Lasso   & $\mathbf{3.4\pm4.1}$ & $\mathbf{3.4\pm3.6}$ & 8.9$\pm$7.2 & $\mathbf{4.9\pm5.7}$ \\*
                       & M-PCA0  & 4.7$\pm$4.9 & 6.3$\pm$5.5 & $\mathbf{3.1\pm4.6}$ & 28.4$\pm$7.7 \\*
                       & M-PCA1a & 4.6$\pm$4.7 & 5.4$\pm$4.9 & 3.4$\pm$4.4 & 21.4$\pm$10.6 \\*
                       & M-PCA1b & 4.9$\pm$5.1 & 5.3$\pm$5.2 & 4.0$\pm$5.2 & 18.6$\pm$9.9 \\*
                       & M-PCA2  & 5.3$\pm$5.0 & 4.4$\pm$5.2 & 3.6$\pm$5.6 & 26.4$\pm$8.2 \\*
  \midrule
  \multirow{7}{*}{28}  & PCA     & 4.7$\pm$4.7 & 5.3$\pm$4.7 & 2.9$\pm$4.6 & 16.3$\pm$9.7 \\*
                       & PLS     & 4.4$\pm$4.5 & 4.4$\pm$4.8 & 3.9$\pm$5.6 & 30.7$\pm$12.2 \\*
                       & Lasso   & $\mathbf{2.9\pm5.0}$ & $\mathbf{3.4\pm4.6}$ & 8.1$\pm$6.9 & $\mathbf{4.1\pm5.4}$ \\*
                       & M-PCA0  & 4.1$\pm$4.4 & 4.6$\pm$4.9 & $\mathbf{2.6\pm4.5}$ & 33.6$\pm$4.4 \\*
                       & M-PCA1a & 4.6$\pm$4.5 & 4.6$\pm$4.7 & 3.1$\pm$4.6 & 25.7$\pm$8.9 \\*
                       & M-PCA1b & 4.6$\pm$4.5 & 4.3$\pm$4.6 & 3.1$\pm$4.6 & 24.7$\pm$8.9 \\*
                       & M-PCA2  & 4.4$\pm$4.8 & 5.1$\pm$5.4 & 3.7$\pm$5.4 & 16.1$\pm$11.7 \\*
  \bottomrule
 \end{tabular}
}
\end{center}
\end{table}
\begin{table}[!ht]
%\captionsetup{width=0.5\textwidth,font=footnotesize}
\caption{Duke Test Errors} \label{table-duke}
\begin{center}
\scalebox{0.9}{
 \begin{tabular}{clcccc}
  \toprule
   K & Scheme & SVM & LR & FLD & NB \\
  \midrule
  \multirow{7}{*}{8}   & PCA     & 14.9$\pm$11.6 & 16.4$\pm$12.1 & 16.2$\pm$11.7 & 22.0$\pm$15.5 \\*
                       & PLS     & $\mathbf{10.7\pm9.2}$ & $\mathbf{11.8\pm10.4}$ & $\mathbf{9.8\pm8.0}$ & 42.9$\pm$13.1 \\*
                       & Lasso   & 18.7$\pm$11.1 & 19.3$\pm$11.6 & 18.0$\pm$12.9 & 20.4$\pm$12.4 \\*
                       & M-PCA0  & 12.9$\pm$9.6 & 13.8$\pm$11.6 & 13.8$\pm$10.9 & $\mathbf{19.8\pm14.8}$ \\*
                       & M-PCA1a & 13.1$\pm$9.2 & 13.3$\pm$11.0 & 12.0$\pm$10.2 & 21.8$\pm$16.0 \\*
                       & M-PCA1b & 13.1$\pm$8.9 & 13.1$\pm$11.6 & 12.4$\pm$10.7 & 20.2$\pm$15.8 \\*
                       & M-PCA2  & 14.2$\pm$9.5 & 12.9$\pm$12.4 & 14.2$\pm$11.5 & 31.6$\pm$15.3 \\*
  \midrule
  \multirow{7}{*}{17}  & PCA     & 13.8$\pm$8.9 & 12.9$\pm$10.8 & 13.1$\pm$11.6 & 27.8$\pm$17.4 \\*
                       & PLS     & $\mathbf{10.9\pm9.4}$ & $\mathbf{11.6\pm9.8}$ & $\mathbf{10.4\pm8.5}$ & 49.8$\pm$12.9 \\*
                       & Lasso   & 16.2$\pm$11.0 & 16.4$\pm$10.8 & 18.0$\pm$10.3 & $\mathbf{19.8\pm12.8}$ \\*
                       & M-PCA0  & 13.1$\pm$8.9 & 12.2$\pm$10.1 & 12.4$\pm$11.1 & 21.8$\pm$13.6 \\*
                       & M-PCA1a & 13.1$\pm$8.9 & 12.4$\pm$9.9 & 12.0$\pm$11.0 & 33.3$\pm$16.9 \\*
                       & M-PCA1b & 13.3$\pm$9.0 & 12.2$\pm$10.1 & 12.9$\pm$11.1 & 27.3$\pm$18.1 \\*
                       & M-PCA2  & 12.7$\pm$9.8 & 12.2$\pm$10.8 & 12.7$\pm$9.5 & 38.9$\pm$16.4 \\*
  \bottomrule
 \end{tabular}
}
\end{center}
\end{table}

\clearpage
\section*{Acknowledgement}
XL is supported by an internal study award at the University of Waikato.
\bibliographystyle{plain}

\end{document}